\documentclass[conference,letterpaper]{IEEEtran}
\IEEEoverridecommandlockouts

\usepackage{cite}
\usepackage{caption}

\usepackage{geometry}
\usepackage{subcaption}
\usepackage{stfloats}
\usepackage{amsmath,amssymb,amsfonts}
\usepackage{graphicx}
\usepackage{textcomp}
\usepackage{xcolor}
\usepackage{tikz}
\usetikzlibrary{decorations.pathreplacing,decorations.markings}
\usepackage{diagbox}
\usepackage{multirow}
\usepackage{makecell}
\usepackage{algorithm}
\usepackage{algpseudocode} 
\usepackage{tikz}
\usetikzlibrary{calc,shadings}
\usepackage{tikz-3dplot}
\usetikzlibrary{arrows.meta, positioning, decorations.pathreplacing}
\def\BibTeX{{\rm B\kern-.05em{\sc i\kern-.025em b}\kern-.08em
   T\kern-.1667em\lower.7ex\hbox{E}\kern-.125emX}}

\newtheorem{definition}{\textbf{Definition}}

\renewcommand{\IEEEbibitemsep}{0pt plus 0.5pt}
\makeatletter
\IEEEtriggercmd{\reset@font\normalfont\fontsize{7.9pt}{8.40pt}\selectfont}
\makeatother
\IEEEtriggeratref{1}

\usepackage[capitalize]{cleveref} 
\usetikzlibrary{calc}

\begin{document}

\newgeometry{top=1in, bottom=0.75in, left=0.75in, right=0.75in}

\title{Trust as a Field: A Macroscopic Representation for Vehicular Networks\\
}

\author{\IEEEauthorblockN{Md Mahmudul Islam}
\IEEEauthorblockA{\textit{Dept. of Civil, Environmental, and Construction Eng.} \\
\textit{University of Central Florida}\\
Orlando, USA \\
mahmudul@ucf.edu}
\and
\IEEEauthorblockN{Shaurya Agarwal}
\IEEEauthorblockA{\textit{Dept. of Civil, Environmental, and Construction Eng.} \\
\textit{University of Central Florida}\\
Orlando, USA \\
shaurya.agarwal@ucf.edu}
}

\IEEEaftertitletext{\vspace{-2\baselineskip}}
\maketitle

\begin{abstract}
Trust assessment is a fundamental component of cooperative and connected vehicle systems. However, existing approaches operate primarily at the level of individual vehicles, making it difficult to reason about trust evolution across road segments. In this paper, we propose a spatio-temporal trust-field framework that aggregates microscopic vehicle-level trust into a continuous representation over space and time. The trust field is formally defined on road segments. We conducted simulation-based experiments using synthetic trajectories generated under controlled conditions, enabling analysis of trust-field behavior in simple road scenarios. Beyond theoretical modeling, we study an implication of the trust-field concept: reconstructing the full trust field from sparse roadside-unit (RSU) measurements. We compare (i) a coordinate-based deep learning baseline that learns a generic trust field from sparse samples and (ii) a field-informed deep learning method that treats trust as a latent quantity carried by vehicles and enforces measurement consistency through the aggregation mechanism. The field-informed approach more accurately recovers trajectory-aligned low-trust patterns and yields improved reconstruction error.
\end{abstract}


\section{Introduction}

Transportation systems continue to struggle with long-standing challenges, including congestion, delays, crashes, and emissions \cite{fhwa}. With the advent of next-generation connected and autonomous vehicles (CAVs), new opportunities have emerged to address these issues through cooperation among road users. Connectivity enables vehicles, cyclists, pedestrians, and infrastructure nodes to share information and coordinate actions in real time, forming the foundation of \emph{cooperative mobility}. Such cooperation can significantly enhance traffic efficiency and safety by enabling human drivers or automated driving features to make timely, informed decisions in response to evolving traffic conditions \cite{alam2022state, thandavarayan2020cooperative}.

Realizing cooperative mobility requires reliable, low-latency communication among mobile agents and efficient decision-making at the node level. However, a vehicle’s willingness to share information, participate in cooperative applications, or act on received information is shaped by its \emph{trust} in the network. This trust encompasses beliefs about the trustworthiness of other agents, the authenticity of information, and the expectation of reciprocal participation \cite{hussain2020trust, hbaieb2022survey}.

Existing trust computation mechanisms in VANETs are broadly categorized as \emph{distributed} or \emph{centralized}. Distributed approaches allow individual nodes to compute trust scores for neighbors based on direct interactions, observations, and message verification~\cite{ltifi2016smart, sedjelmaci2015accurate, wahab2014cooperative}, whereas centralized frameworks rely on trusted authorities or RSUs to manage reputation information and certificate validation~\cite{li2012reputation, li2013rgte, bissmeyer2012central}. Trust evaluation methodologies also differ conceptually: \emph{entity-centric} models assess the trustworthiness of agents based on historical behavior and cooperation patterns~\cite{marmol2012trip}, while \emph{data-centric} schemes evaluate the reliability of received messages through consistency and redundancy checks~\cite{hussain2016hybrid, rawat2015trust}. Game-theoretic formulations further model trust as an outcome of strategic interactions among rational agents~\cite{shivshankar2015evolutionary}.

To capture trust dynamics over time, researchers have proposed Bayesian updating schemes, fuzzy-logic aggregation methods, and statistical or machine-learning models such as logistic regression to estimate evolving trust trends~\cite{pham2018adaptive, hu2016replace, wang2016catrust}. Building on these foundations, broader \emph{trust management} systems integrate prediction, malicious-node detection, and cooperation incentives through centralized or decentralized architectures, often incorporating cryptographic authentication, consensus mechanisms, or learning-based techniques~\cite{kerrache2016trust, chen2016game}. While these approaches provide rich mathematical and algorithmic tools for node- or link-level trust assessment, they remain fundamentally localized and do not address whether trust can aggregate into structured, macroscopic patterns over road segments or across the transportation network.

In contrast, macroscopic traffic flow theory models aggregate variables such as density and flow as spatio-temporal fields governed by hydrodynamic conservation laws, enabling the study of emergent phenomena such as shockwaves and bottlenecks \cite{lighthill1955kinematic}. This perspective motivates a new representation of trust: instead of treating trust solely as a local node-level quantity, we model it as a \emph{spatio-temporal trust field} defined over road segments. Analogous to congestion or travel-time fields, the proposed trust field assigns a value to each point in space and time, revealing how trust evolves under vehicle motion and traffic dynamics. Such a formulation enables network-level analysis, reduces the need for repeated local trust computations, and supports reconstruction and prediction of trust from sparse measurements.

\restoregeometry

\textbf{Contributions:} The primary contributions of this paper are summarized as follows:

\textbf{(a) Trust-field formulation:} We introduce a mathematical framework for representing trust as a spatio-temporal field over road segments using the aggregation of vehicle-level trust trajectories.

\textbf{(b) Simulation-based validation:} Through controlled microscopic simulations using synthetic trajectories, we demonstrate how vehicle-level trust dynamics give rise to structured macroscopic trust-field patterns. When congestion and shockwave formation are introduced, the trust field reveals how a slow or stopped malicious vehicle creates a localized trust degradation whose spatial extent is shaped by traffic dynamics.

\textbf{Outline:} Section~\ref{sec:trust_field} establishes the proposed spatio-temporal trust-field framework. Section~\ref{sec:sim_env_exp_setup} describes the simulation environment and experimental setup. Section~\ref{sec:experimental_findings} presents empirical results. Section~\ref{sec:implication} discusses reconstruction implications. Finally, Section~\ref{sec:conclusion} concludes the paper.

\section{Establishing the Trust Field} \label{sec:trust_field}
This section introduces the proposed trust-field formulation that maps microscopic vehicle-level trust to a spatio-temporal field representation over road segments. We first outline the underlying assumptions and notation, and then formally define the trust field on the road segments.
\subsection{Assumptions and Vehicle Trust}

\paragraph{Trust of a vehicle}

In the context of vehicular systems, we interpret a vehicle's trust value $\tau_i(t) \in [0,1]$ as an externally assigned scalar representing the degree of confidence in the expected future behavior of vehicle $i$ at time $t$. A value of $\tau_i(t)=1$ corresponds to fully trusted behavior, while a value of $\tau_i(t)=0$ indicates no trust. Intermediate values capture varying degrees of uncertainty or risk attributed to the vehicle's future actions.

\paragraph{Assumptions}
Throughout this work we assume that the trust score of every vehicle in the traffic stream is known at any time $t \in [0,T_{\max}]$. The mechanism by which these trust scores are computed is deliberately left unspecified. We treat trust as an externally supplied quantity that may arise from any trust model. Our goal is not to compute trust but rather to \emph{represent, propagate, and analyze} trust as a spatio-temporal field over a road segment.

\subsection{Trust Field on a Road Segment} \label{sec:TF_Road}

Before formalizing the trust field mathematically, we first provide an intuitive illustration of the concept. As vehicles move and interact, trust variations are not static or localized, but propagate through space and time in a structured manner.

\begin{figure}[ht!]
    \centering
    \includegraphics[width=0.9\linewidth]{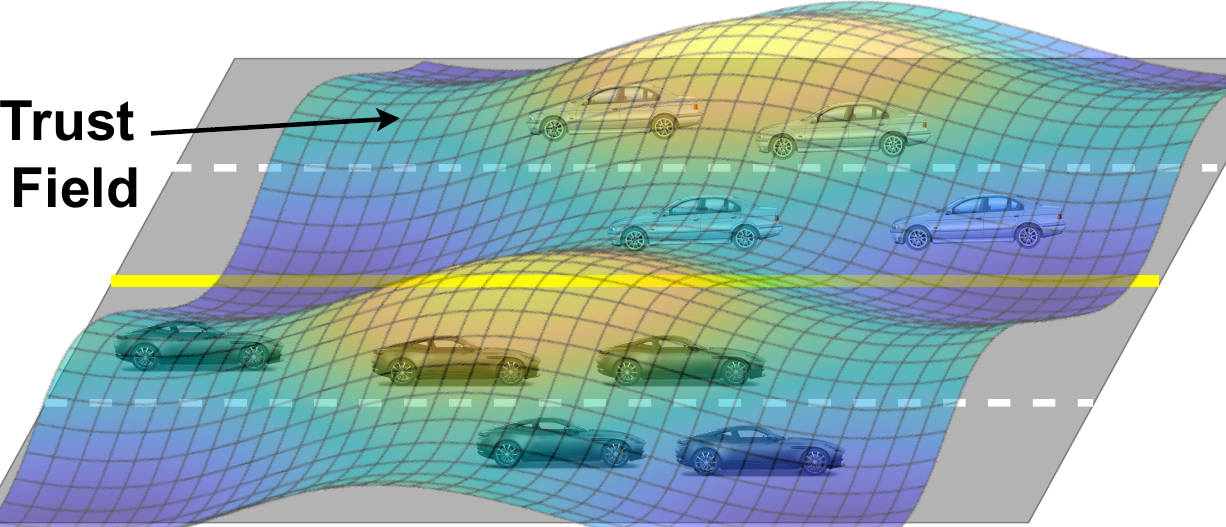}
    \caption{Concept of Trust Field.}
    \label{fig:trust_field_concept}
\end{figure}

Figure~\ref{fig:trust_field_concept} depicts this idea, where trust is represented as a continuous spatio-temporal surface shaped by vehicle motion and interactions. This conceptual view motivates modeling trust as a field quantity rather than as isolated scalar values associated with individual vehicles or messages.

Motivated by this, we define the \textit{Trust Field} for a single road segment as a scalar spatiotemporal function that represents the degree of reliability or confidence within a vehicular network along the length of the road segment. 
\begin{definition}[\textbf{Trust Field on a Road Segment}]
Let the road segment be represented as a one-dimensional spatial interval $x \in [0, L]$, where $x = 0$ denotes the beginning of the road segment, and $x = L$ denotes the end of the road segment. Let time be denoted by $t \in \mathbb{R}$. Then the \emph{trust field} for the road segment is defined as the function
\begin{equation}
    T : [0, L] \times \mathbb{R^+} \to [0,1]
\end{equation}
which assigns to each spatio-temporal point $(x,t)$ a scalar trust value $T(x,t)$ in the range $[0,1]$. Here, \(T = 1\) indicates complete trustworthiness, whereas \(T = 0\) corresponds to complete distrust.
\end{definition}

\begin{definition}[\textbf{Trust Trajectory}]
For a vehicle traveling along a road segment, we define its \emph{trust trajectory} as a function $\tau_v : \mathbb{R} \to [0,1]$ given by
\begin{equation}
    \tau_v(t) = T\bigl(x_v(t),\, t\bigr),
\end{equation}
where $x_v(t)$ denotes the position of the vehicle $v$ at time $t$. The function $\tau_v(t)$ represents the trust value associated with the vehicle $v$ at position $x_v(t)$ and time $t$, thereby capturing the evolution of trust along the vehicle’s motion.
\end{definition}

The collection of such trust trajectories across all vehicles induces a microscopic description of trust that is inherently discrete in space. To obtain a macroscopic, spatially continuous representation, these vehicle-level trust trajectories must be aggregated into a spatio-temporal trust field.

\paragraph{Aggregation Framework}
Let $N(t)$ denote the number of active vehicles at time $t$, with positions $\{x_i(t)\}_{i=1}^{N(t)}$ and corresponding trust values $\{\tau_i(t)\}_{i=1}^{N(t)}$. We define the macroscopic trust field $T(x,t)$ as a normalized spatial aggregation of these microscopic trust values:
\begin{equation}
    T(x,t)
    =
    \mathcal{S}\!\left(
    \{\tau_i(t)\}, \{x_i(t)\};\, x
    \right),
\end{equation}
where $\mathcal{S}(\cdot)$ denotes an aggregation operator that maps discrete trust samples onto a continuous spatial domain. The specific choice of aggregation operator is not intrinsic to the trust field concept and may vary depending on modeling objectives, resolution requirements, or computational considerations.

\paragraph{Gaussian Kernel Instantiation}
In this work, we employ a kernel-based aggregation scheme as a concrete instantiation of the above framework. Specifically, we use a Gaussian kernel to construct a smooth spatial trust field:
\begin{equation}
T(x,t) =
\frac{\displaystyle \sum_{i=1}^{N(t)} \tau_i(t)\,
K_\sigma\!\big(x - x_i(t)\big)}
{\displaystyle \sum_{i=1}^{N(t)} K_\sigma\!\big(x - x_i(t)\big)},
\label{eq:trust_field}
\end{equation}
where $K_\sigma(\Delta x) = \exp\!\left[-\frac{(\Delta x)^2}{2\sigma^2}\right]$
is a Gaussian kernel with bandwidth $\sigma$ controlling the spatial influence of individual vehicles.

This kernel-based formulation provides a smooth and numerically stable mapping from discrete vehicle-level trust values to a continuous trust field. We emphasize, however, that alternative aggregation strategies (e.g., binning, compact-support kernels, or other weighting schemes) could be adopted without altering the fundamental notion of a macroscopic trust field \cite{kachroo2023nonlocal, avila2020data}.

\section{Simulation Environment and Experimental Setup}
\label{sec:sim_env_exp_setup}

This section describes the simulation framework and experimental design used to validate the proposed spatio-temporal trust-field formulation. 

\subsection{Simulation Environment}

In this paper, we use synthetic vehicle trajectories generated in a controlled simulation environment. By using synthetic trajectories, we retain full control over vehicle interactions, routing behavior, and traffic demand, thereby enabling a systematic validation of the trust-field formulation in minimal settings. Once the correctness and interpretability of the trust-field representation are established in these controlled environments, the framework can be naturally extended to real-world trajectory datasets as part of future work.

\begin{table}[ht!]
\centering
\caption{IDM parameter values used in the simulations.}
\label{tab:idm_params}
\begin{tabular}{lll}
\hline
\textbf{Description} & \textbf{Value (with unit)} \\ \hline
Desired speed of vehicle $i$ 
           & $\approx 30$--$31.5~\text{m/s}$ \\
Maximum acceleration 
           & $1.5~\text{m/s}^2$ \\
Comfortable deceleration 
           & $2.0~\text{m/s}^2$ \\
Desired time headway 
           & $1.2~\text{s}$ \\
Minimum gap (jam distance) 
           & $2~\text{m}$ \\
IDM acceleration exponent 
           & $4$ \\
Vehicle length 
           & $4.5~\text{m}$ \\ \hline
\end{tabular}
\end{table}

All synthetic vehicle trajectories are generated using a microscopic traffic simulator implemented in \textsc{MATLAB}. Vehicle longitudinal motion on each road segment follows the Intelligent Driver Model (IDM) \cite{treiber2000congested}, a widely adopted car-following model in traffic-flow studies. A common set of IDM parameters \cite{albeaik2022limitations} is used across all experiments, as summarized in \cref{tab:idm_params}.

The vehicle trajectories contains positions and velocities of the vehicles, which serve as inputs to the trust-field construction pipeline.

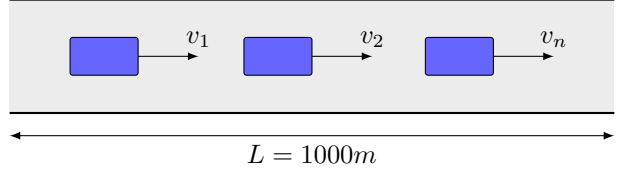
\begin{figure}[ht!]
\centering
\begin{tikzpicture}[scale=1.0, >=latex]
  \fill[gray!15] (0,0) rectangle (8,1.5);
  \draw[thick] (0,0) -- (8,0);
  \draw[thick] (0,1.5) -- (8,1.5);
  
  \draw[<->] (0,-0.3) -- (8,-0.3)
    node[midway, below] {$L = 1000m$};
  
  \draw[fill=blue!60, rounded corners=1pt] (0.8,0.5) rectangle ++(0.9,0.5);
  \node at (1.7,0.75) (c1) {};
  \draw[-latex] (c1.center)--+(0.8,0) node[above] {$v_1$};

  \draw[fill=blue!60, rounded corners=1pt] (3.1,0.5) rectangle ++(0.9,0.5);
  \node at (4.0,0.75) (c1) {};
  \draw[-latex] (c1.center)--+(0.8,0) node[above] {$v_2$};

  \draw[fill=blue!60, rounded corners=1pt] (5.5,0.5) rectangle ++(0.9,0.5);
  \node at (6.4,0.75) (c1) {};
  \draw[-latex] (c1.center)--+(0.8,0) node[above] {$v_n$};

\end{tikzpicture}
\caption{Single lane road network of length $L = 1000m$.}
\label{fig:single_lane_road}
\end{figure}

\subsection{Experimental Setup}

\subsubsection{Single-Lane Road Segment}

We first consider a single-lane road segment of length $L = 1000~\text{m}$, shown in \cref{fig:single_lane_road}. The spatial domain is represented by the interval $x \in [0,1000]$, and the simulation is run over the time horizon $t \in [0,70]~\text{s}$. A total of $n = 35$ vehicles are initially placed along the segment with feasible headways satisfying the IDM gap constraints.

Vehicles evolve according to the IDM dynamics, resulting in continuous position trajectories along with corresponding speed profiles and inter-vehicle spacings. These data are used to construct the spatio-temporal trust field on the road segment using the method described in \cref{sec:TF_Road}. This scenario serves as a baseline setting for examining trust-field formation in a one-dimensional environment.

\begin{figure*}[htbp]
    \centering

    \begin{subfigure}[b]{0.3\textwidth}
        \centering
        \includegraphics[width=\textwidth]{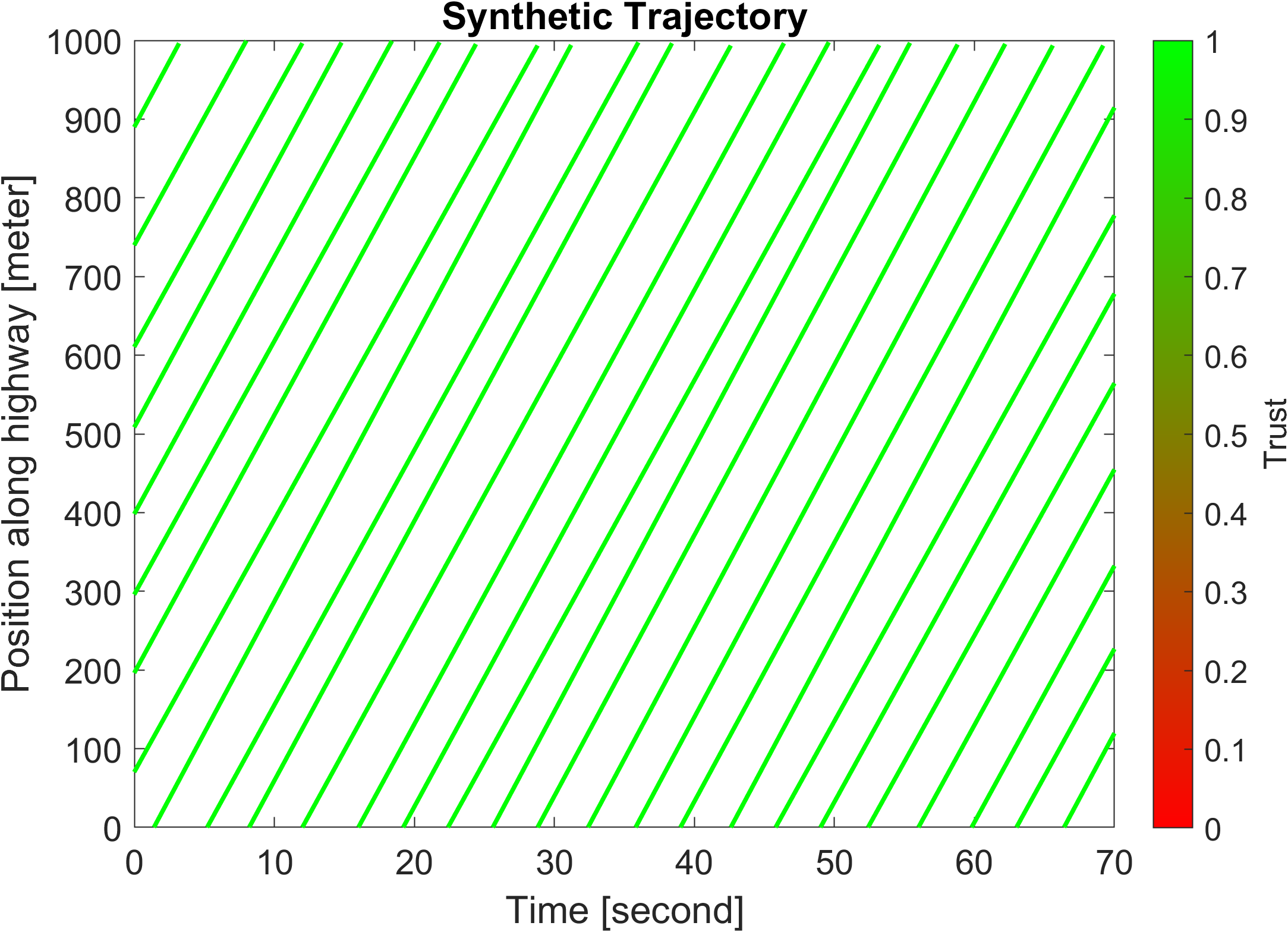}
        \caption{Baseline trust trajectory}
        \label{fig:sub1}
    \end{subfigure}
    \hfill
    \begin{subfigure}[b]{0.3\textwidth}
        \centering
        \includegraphics[width=\textwidth]{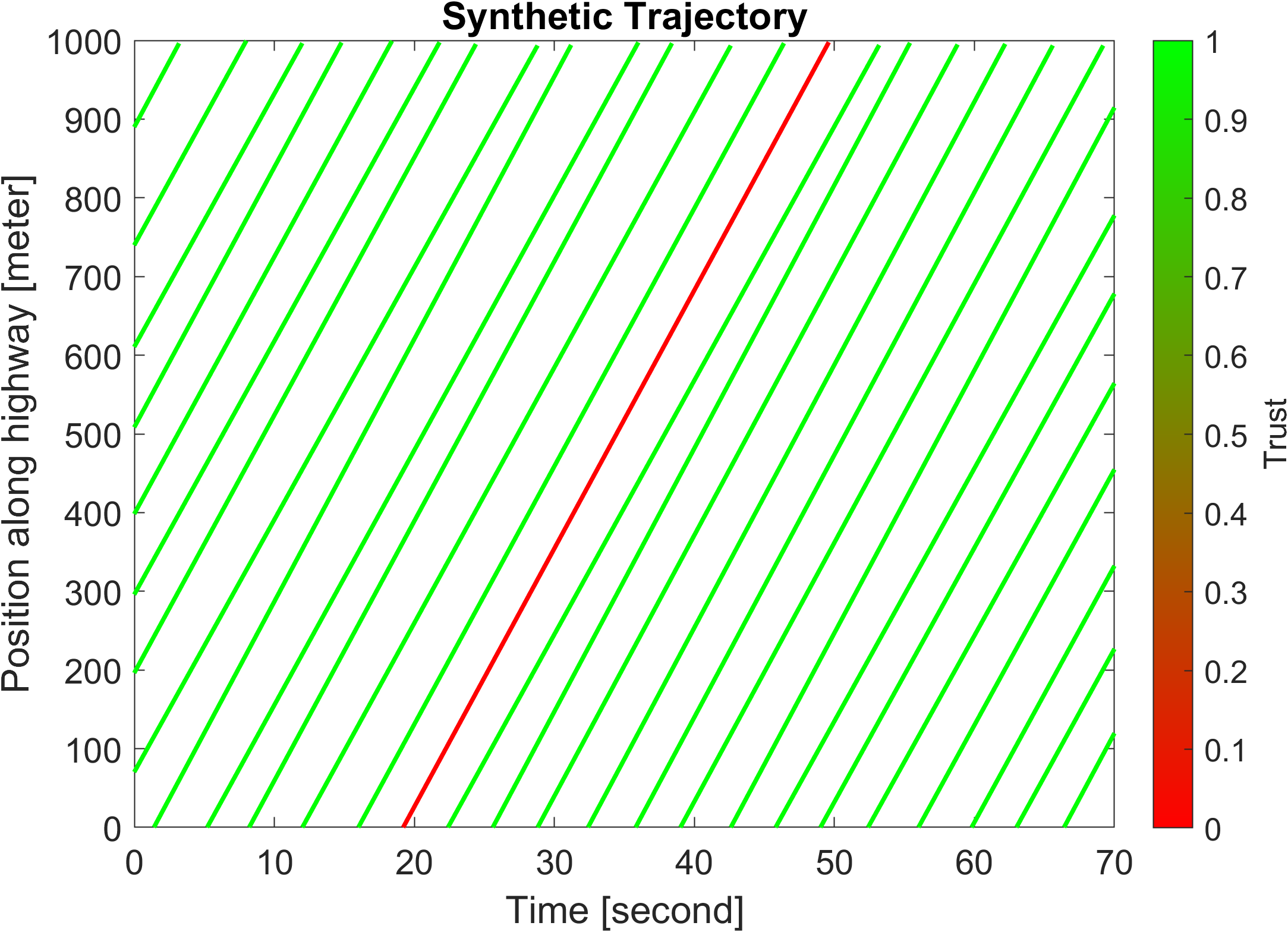}
        \caption{Static trust trajectory}
        \label{fig:sub2}
    \end{subfigure}
    \hfill
    \begin{subfigure}[b]{0.3\textwidth}
        \centering
        \includegraphics[width=\textwidth]{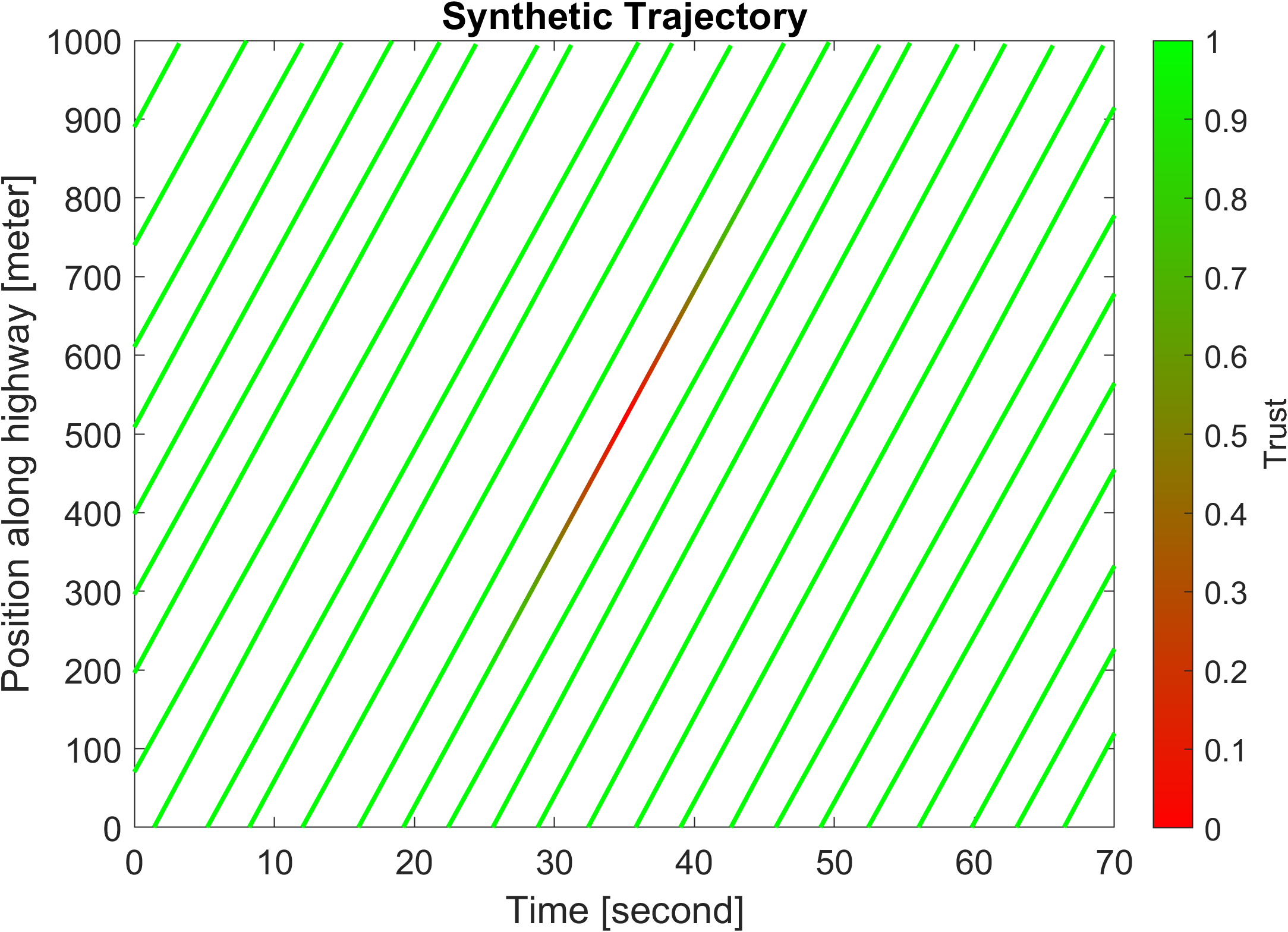}
        \caption{Dynamic trust trajectory}
        \label{fig:sub3}
    \end{subfigure}


    \begin{subfigure}[b]{0.3\textwidth}
        \centering
        \includegraphics[width=\textwidth]{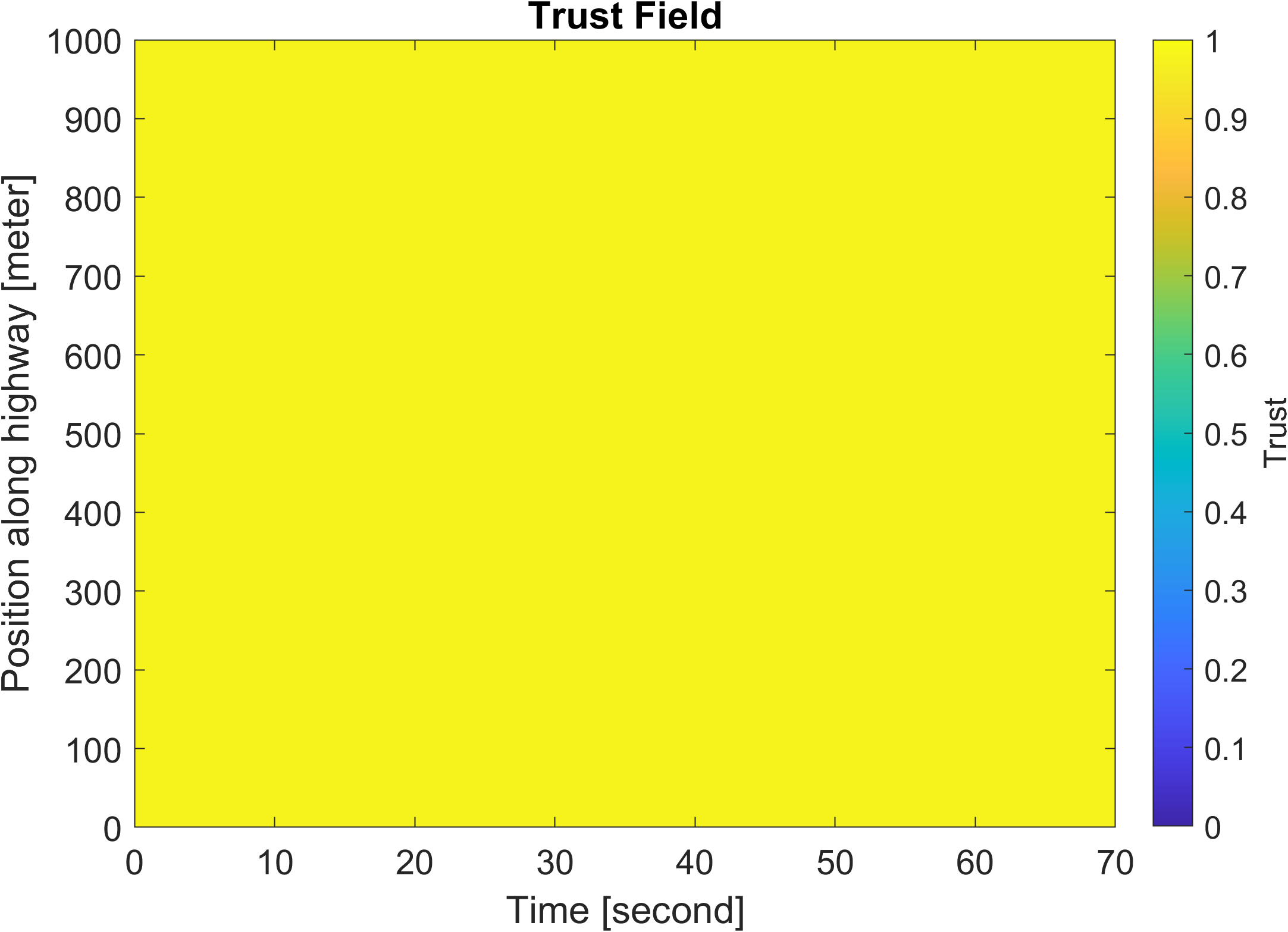}
        \caption{Baseline trust field}
        \label{fig:sub4}
    \end{subfigure}
    \hfill
    \begin{subfigure}[b]{0.3\textwidth}
        \centering
        \includegraphics[width=\textwidth]{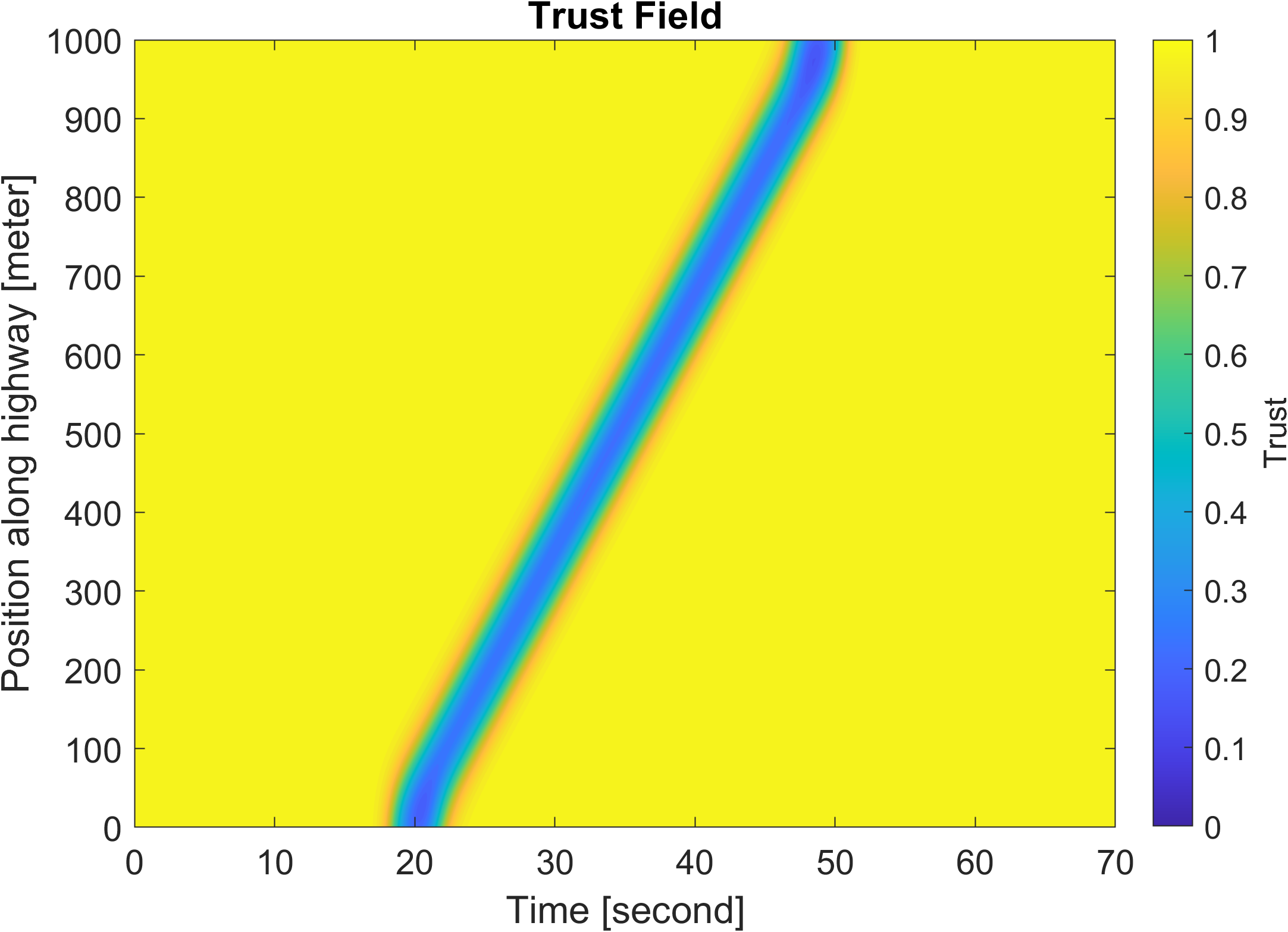}
        \caption{Static trust field}
        \label{fig:sub5}
    \end{subfigure}
    \hfill
    \begin{subfigure}[b]{0.3\textwidth}
        \centering
        \includegraphics[width=\textwidth]{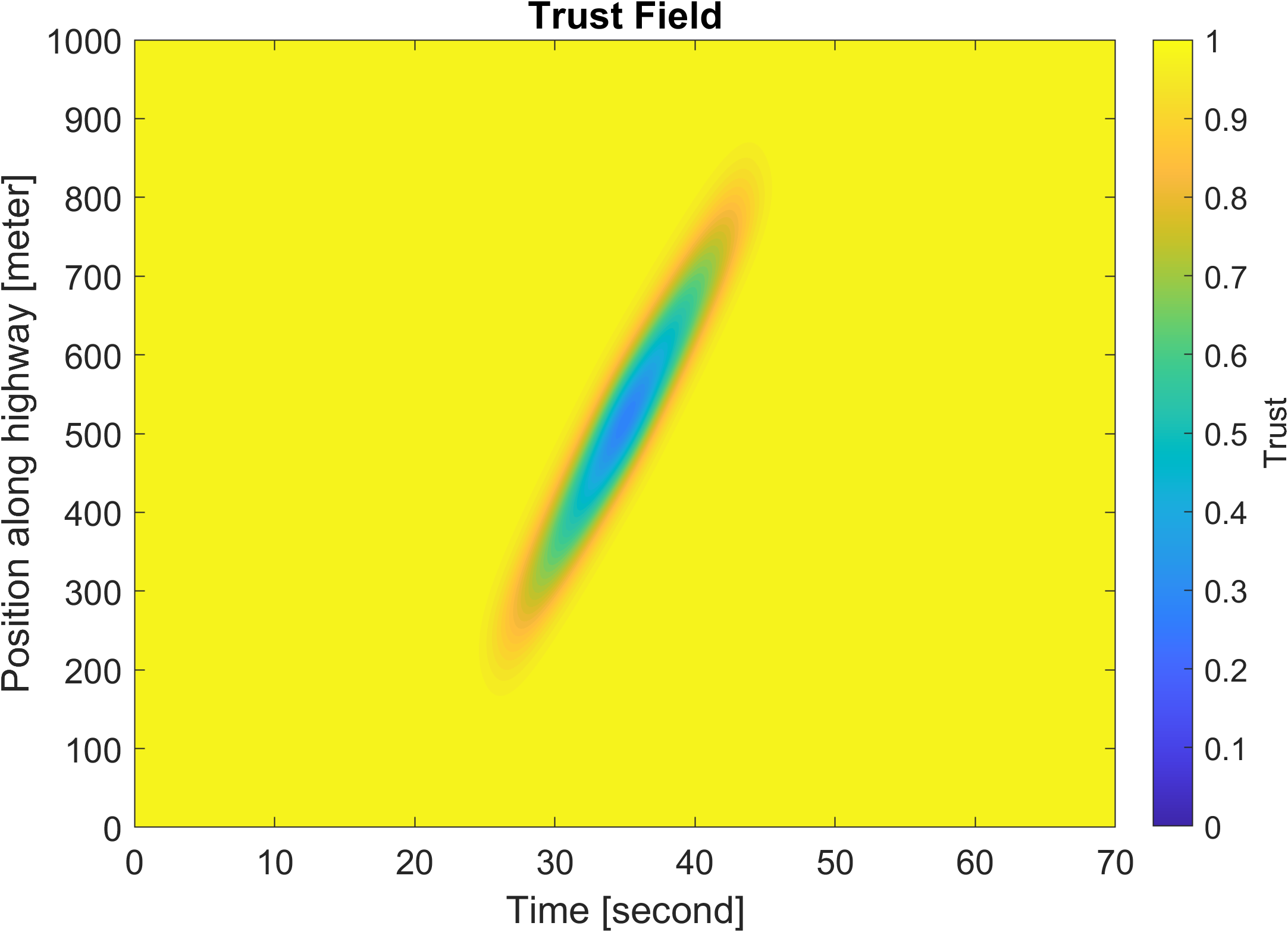}
        \caption{Dynamic trust field}
        \label{fig:sub6}
    \end{subfigure}

    \caption{Trust trajectory and field in a free-flow road segment.}
    \label{fig:freeflow_results}
\end{figure*}

\section{Experimental Findings}
\label{sec:experimental_findings}

In this section, we empirically examine how the proposed trust field behaves under different traffic regimes and attacker behaviors. In all cases, vehicle–level trust trajectories $\{\tau_i(t)\}$ are aggregated into a macroscopic trust field $T(x,t)$ using the Gaussian kernel construction in \eqref{eq:trust_field}.

\subsection{Trust Trajectory and Field in a Free-Flow Road Segment}

Figure~\ref{fig:freeflow_results} illustrates the trust-field behavior on a single-lane free-flow road segment. Vehicle trajectories (top row) are approximately parallel, reflecting homogeneous motion, while the bottom row shows the corresponding macroscopic trust field.

In the baseline case, all vehicles maintain full trust, $\tau_i(t)\equiv 1$ (Fig.~\ref{fig:freeflow_results}(a)), producing a spatially and temporally uniform field $T(x,t)\equiv 1$, shown as a homogeneous region in Fig.~\ref{fig:freeflow_results}(d). This confirms that a fully trustworthy traffic stream yields a constant macroscopic trust field independent of vehicle configuration. We used two types of trust dynamics for the malicious vehicle: static and dynamic

First, a single malicious vehicle with constant low trust is introduced (represented by the red color in Fig.~\ref{fig:freeflow_results}(b)). The Gaussian aggregation generates a narrow diagonal band of low trust aligned with its trajectory (Fig.~\ref{fig:freeflow_results}(e)), while the rest of the domain remains near $T(x,t)=1$. Second, we used dynamic trust for the malicious vehicle, allowing the trust to decrease from $1$ to $0$ and then recover (Fig.~\ref{fig:freeflow_results}(c)). This produces a comet-like low-trust structure (Fig.~\ref{fig:freeflow_results}(f)), whose width and intensity reflect the temporal evolution of $\tau_v(t)$.

Across all cases, the macroscopic trust field closely mirrors vehicle-level trust dynamics: uniform trust yields a flat field, a static attacker creates a thin diagonal low-trust band, and a time-varying attacker generates a structured region encoding its trajectory and trust evolution.

\begin{figure*}[ht!]
    \centering

    \begin{subfigure}[b]{0.3\textwidth}
        \centering
        \includegraphics[width=\textwidth]{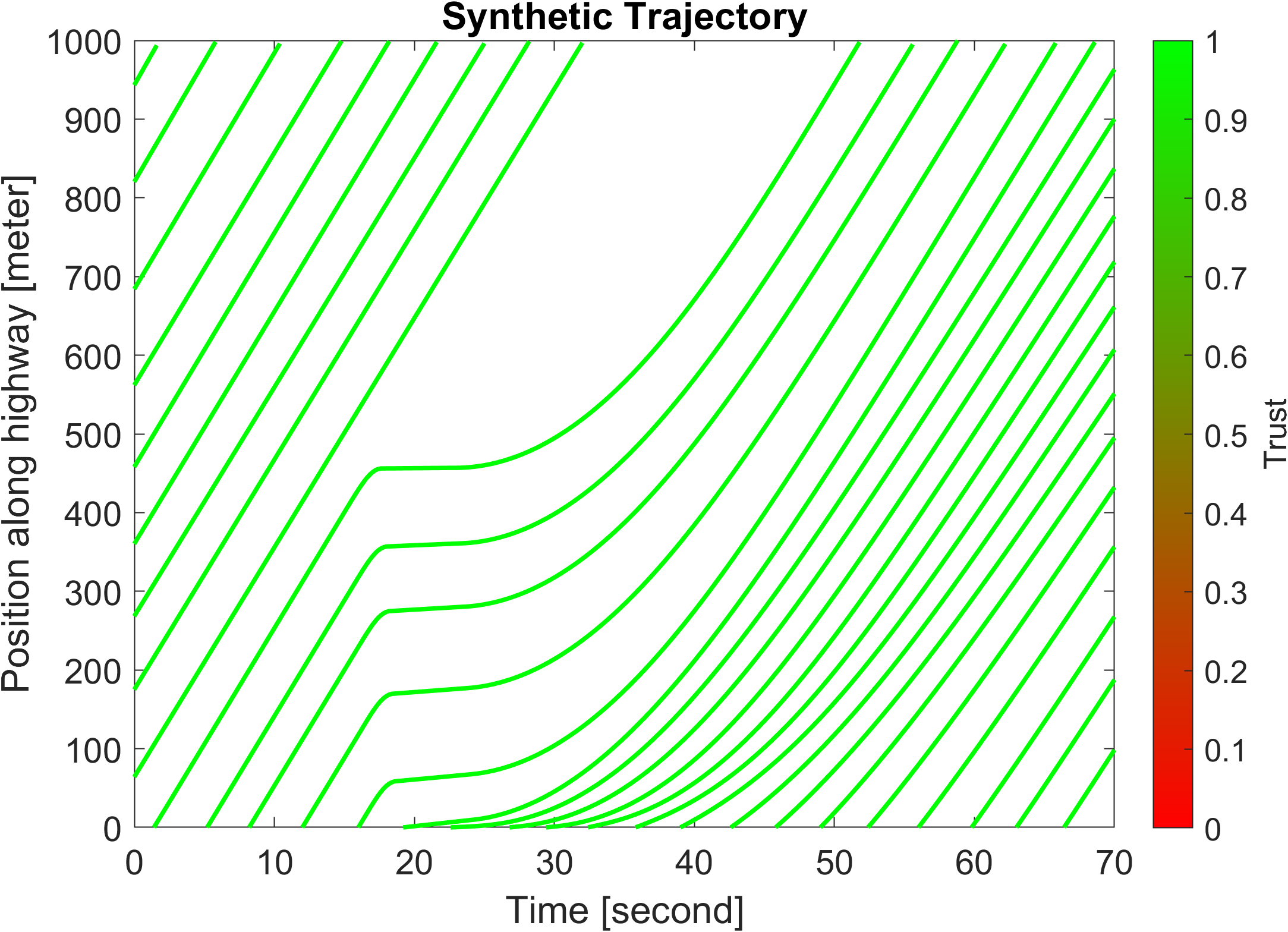}
        \caption{Baseline trust trajectory}
        \label{fig:jam_results_btt}
    \end{subfigure}
    \hfill
    \begin{subfigure}[b]{0.3\textwidth}
        \centering
        \includegraphics[width=\textwidth]{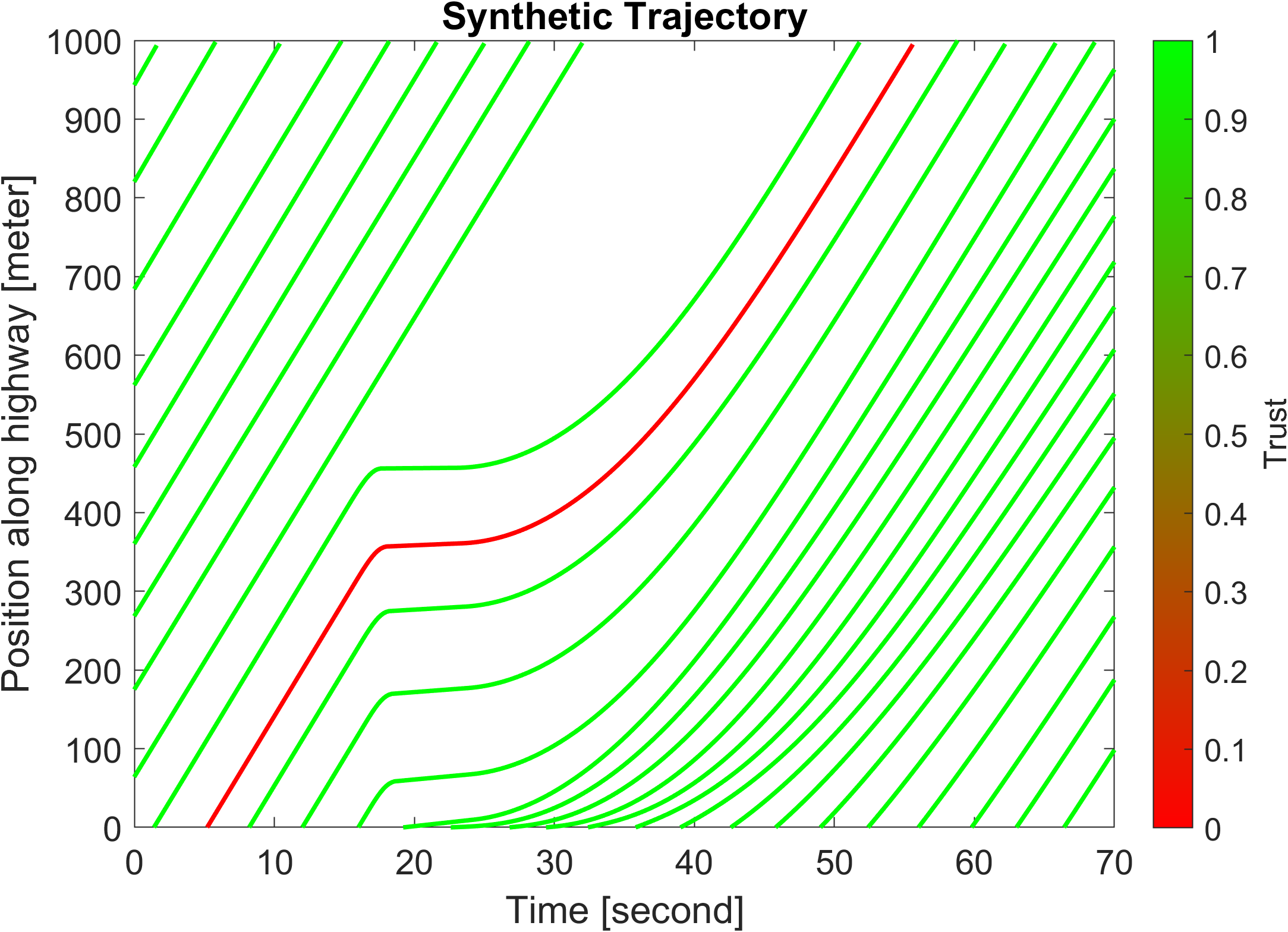}
        \caption{Static trust trajectory}
        \label{fig:jam_results_stt}
    \end{subfigure}
    \hfill
    \begin{subfigure}[b]{0.3\textwidth}
        \centering
        \includegraphics[width=\textwidth]{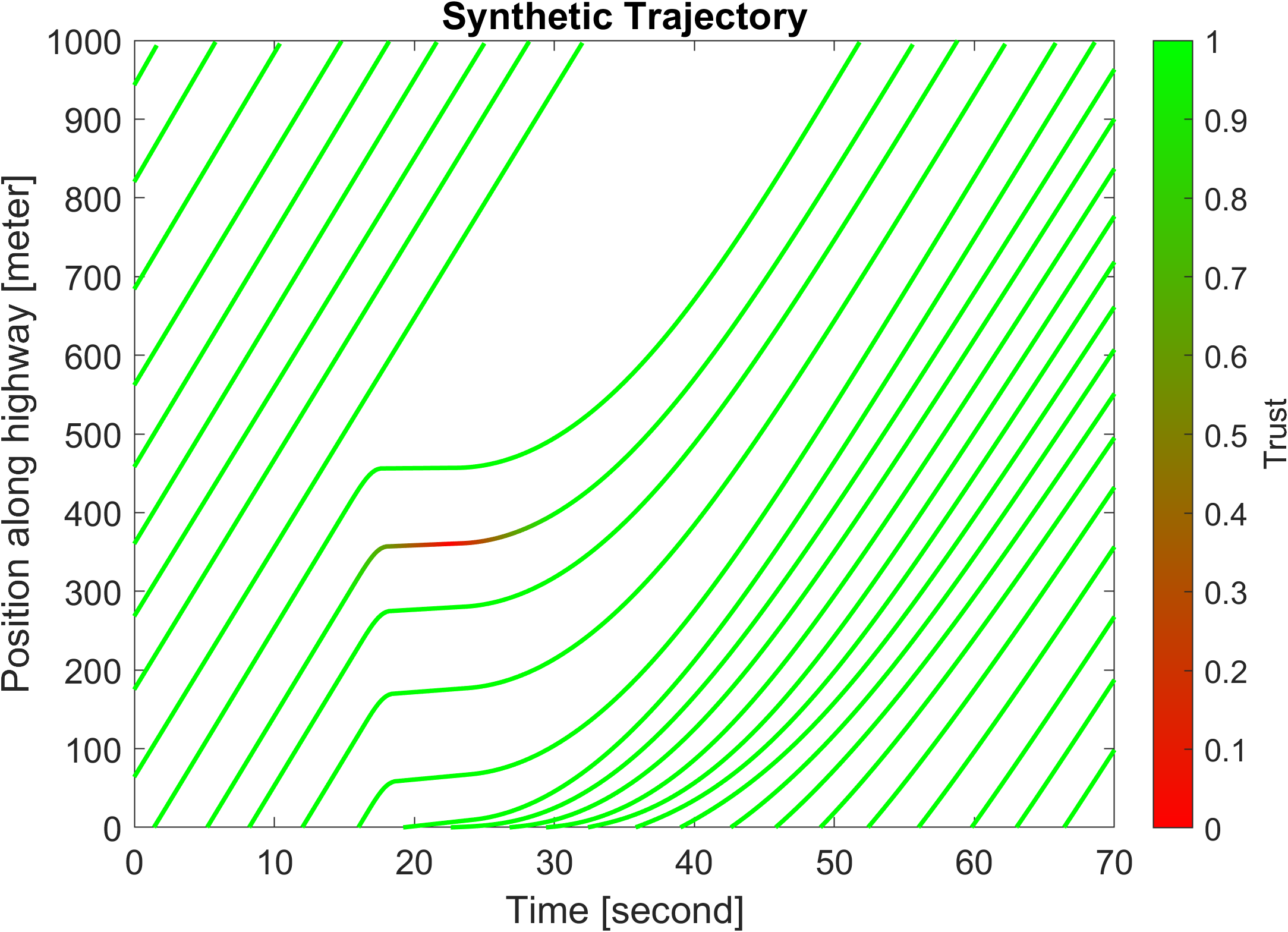}
        \caption{Dynamic trust trajectory}
        \label{fig:jam_results_dtt}
    \end{subfigure}


    \begin{subfigure}[b]{0.3\textwidth}
        \centering
        \includegraphics[width=\textwidth]{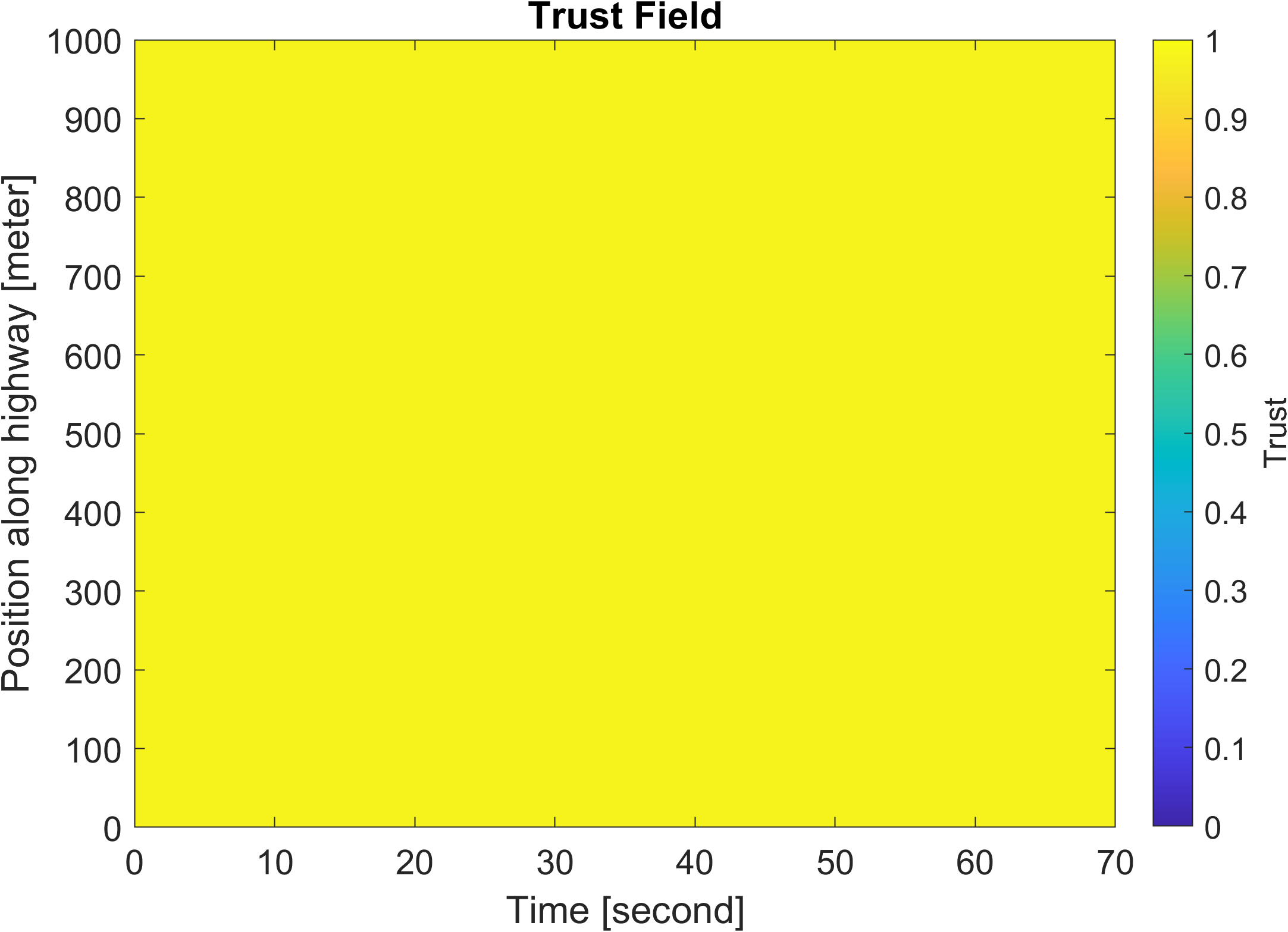}
        \caption{Baseline trust field}
        \label{fig:jam_results_btf}
    \end{subfigure}
    \hfill
    \begin{subfigure}[b]{0.3\textwidth}
        \centering
        \includegraphics[width=\textwidth]{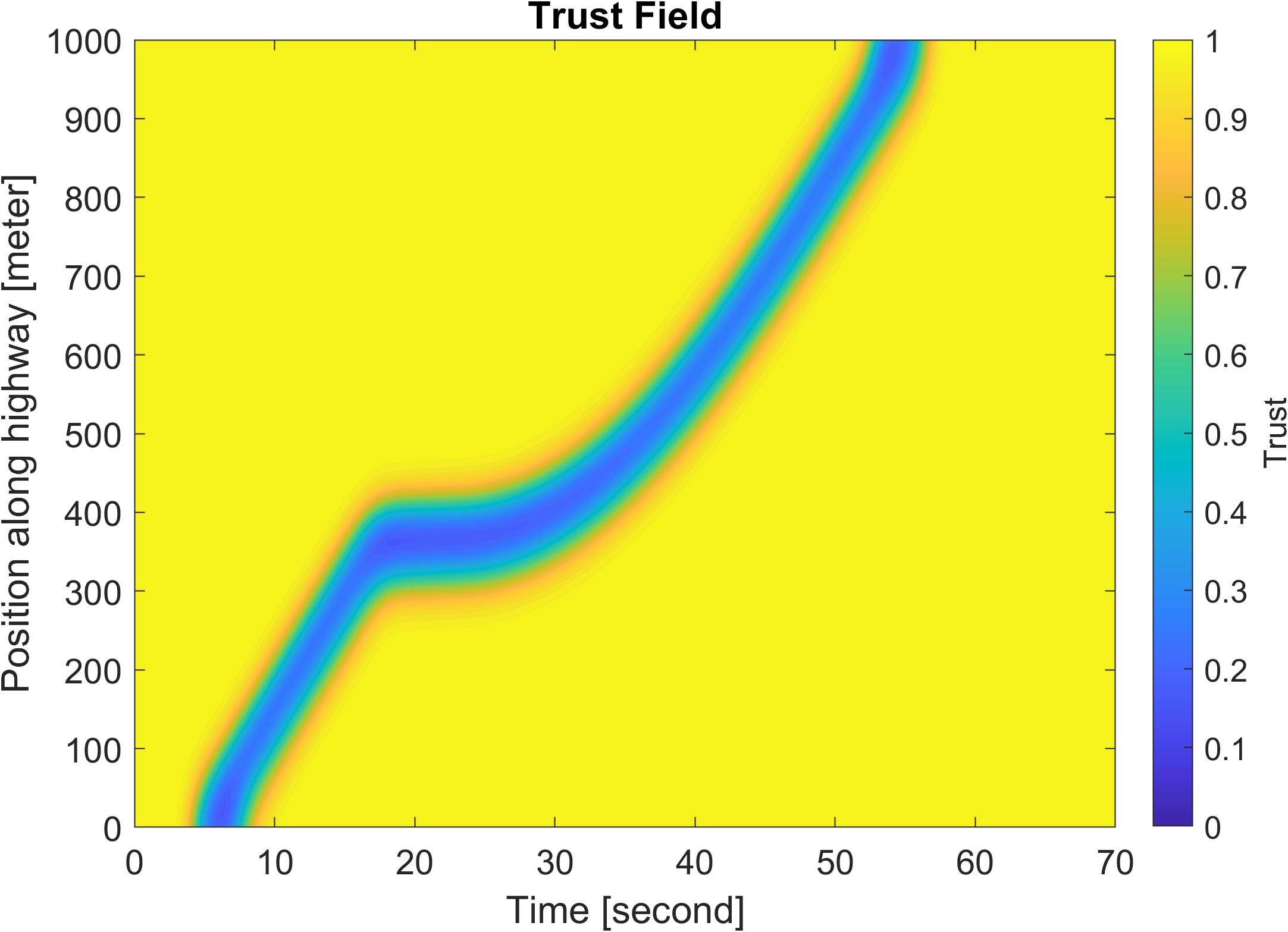}
        \caption{Static trust field}
        \label{fig:jam_results_stf}
    \end{subfigure}
    \hfill
    \begin{subfigure}[b]{0.3\textwidth}
        \centering
        \includegraphics[width=\textwidth]{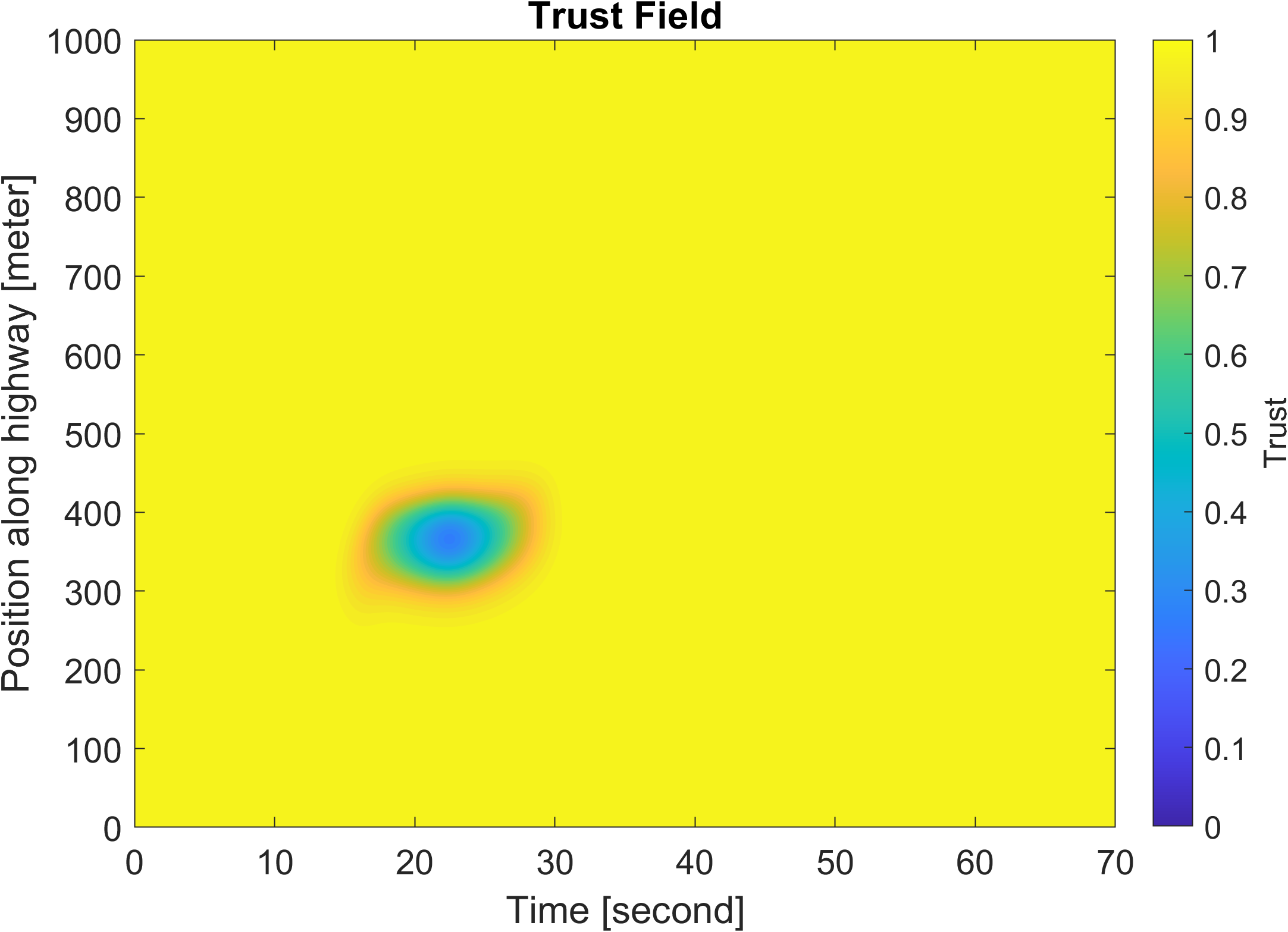}
        \caption{Dynamic trust field}
        \label{fig:jam_results_dtf}
    \end{subfigure}

    \caption{Trust trajectory and field in a road segment where shockwave is present.}
    \label{fig:jam_results}
\end{figure*}

\subsection{Trust Trajectory and Field in a Shockwave Road Segment}

In the second scenario, we impose a traffic jam on the same single-lane road segment to examine the interaction between congestion and trust dynamics (Fig.~\ref{fig:jam_results}). In the baseline case, all vehicles remain fully trusted, $\tau_i(t)\equiv 1$, and although the trajectories in Fig.~\ref{fig:jam_results}(a) exhibit clear congestion with dense clustering, the trust field in Fig.~\ref{fig:jam_results}(d) remains uniformly high. Thus, congestion alone does not distort the macroscopic trust field.

Introducing a single malicious vehicle with a constant low trust value (Fig.~\ref{fig:jam_results}(b)) produces a low-trust region concentrated around the jam (Fig.~\ref{fig:jam_results}(e)). Unlike free flow, the attacker’s prolonged presence in a confined spatial region causes the Gaussian aggregation to accumulate a persistent and spatially extended low-trust band. When the attacker’s trust varies dynamically (Fig.~\ref{fig:jam_results}(c)), the resulting field (Fig.~\ref{fig:jam_results}(f)) shows a widened and intensified low-trust region centered at the congestion location. Because the vehicle remains in a limited area for an extended time, its impact on the trust field is amplified, demonstrating that stationary or slow-moving attackers can strongly depress trust in congested regions despite limited spatial coverage.

\section{Implications of the Trust Field Representation}
\label{sec:implication}

This section examines the practical implications of modeling vehicular trust as a spatio–temporal field rather than as a simple scalar value. In real vehicular networks, trust information is only partially observable due to sparse deployment of trusted RSUs and limited communication ranges. Vehicles must therefore estimate trustworthiness beyond their local sensing range, making trust state estimation analogous to traffic state estimation from limited sensor data. The key challenge is reconstructing the full trust field from sparse observations, particularly when field properties are ignored.

In this experiment, a synthetic trust field is generated over a 1000 m road segment for 240 seconds using the method described in \cref{sec:sim_env_exp_setup}. Trusted RSUs are placed at fixed locations (0, 200, 400, 600, 800, 1000 m), each providing measurements at 240 time points, for a total of 1440 possible observations. To reflect practical constraints, only 70\% of these samples (1008 observations) are used for training. The goal is to estimate the complete spatio–temporal trust field from these sparse data.

To illustrate the implications of trust-field modeling, we compare two trust state estimation strategies. The first uses a deep learning model trained solely on observed trust samples. The second adopts a field-informed approach that models trust as a latent variable carried by vehicles and incorporates trajectory and spatio–temporal structure. The comparison demonstrates the importance of trust field modeling for reliable estimation under sparse sensing conditions.

\begin{figure*}[t!]
    \centering

    \begin{subfigure}[b]{0.32\textwidth}
        \centering
        \includegraphics[width=\textwidth]{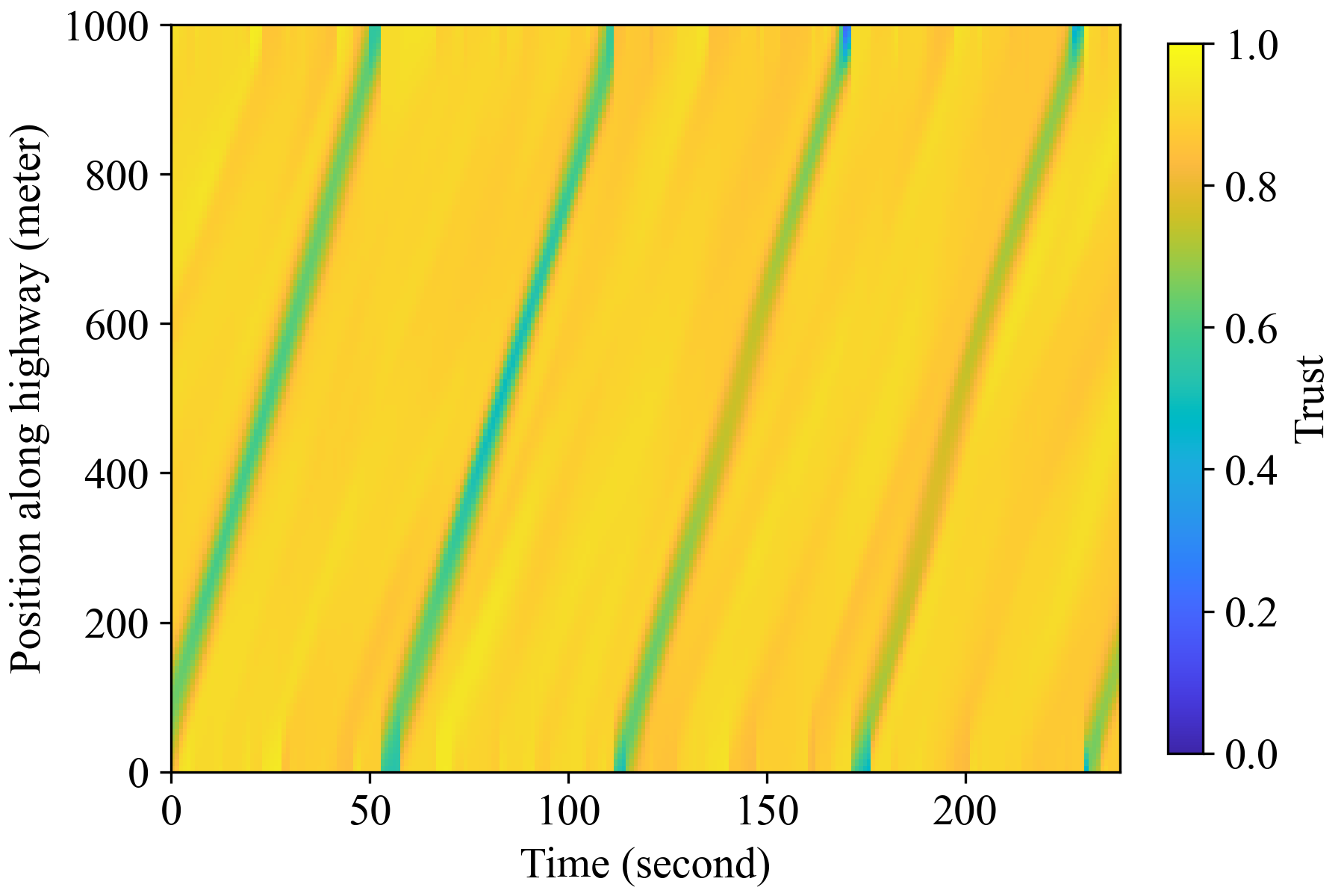}
        \caption{Ground truth}
        \label{fig:impl_ground}
    \end{subfigure}
    \hfill
    \begin{subfigure}[b]{0.32\textwidth}
        \centering
        \includegraphics[width=\textwidth]{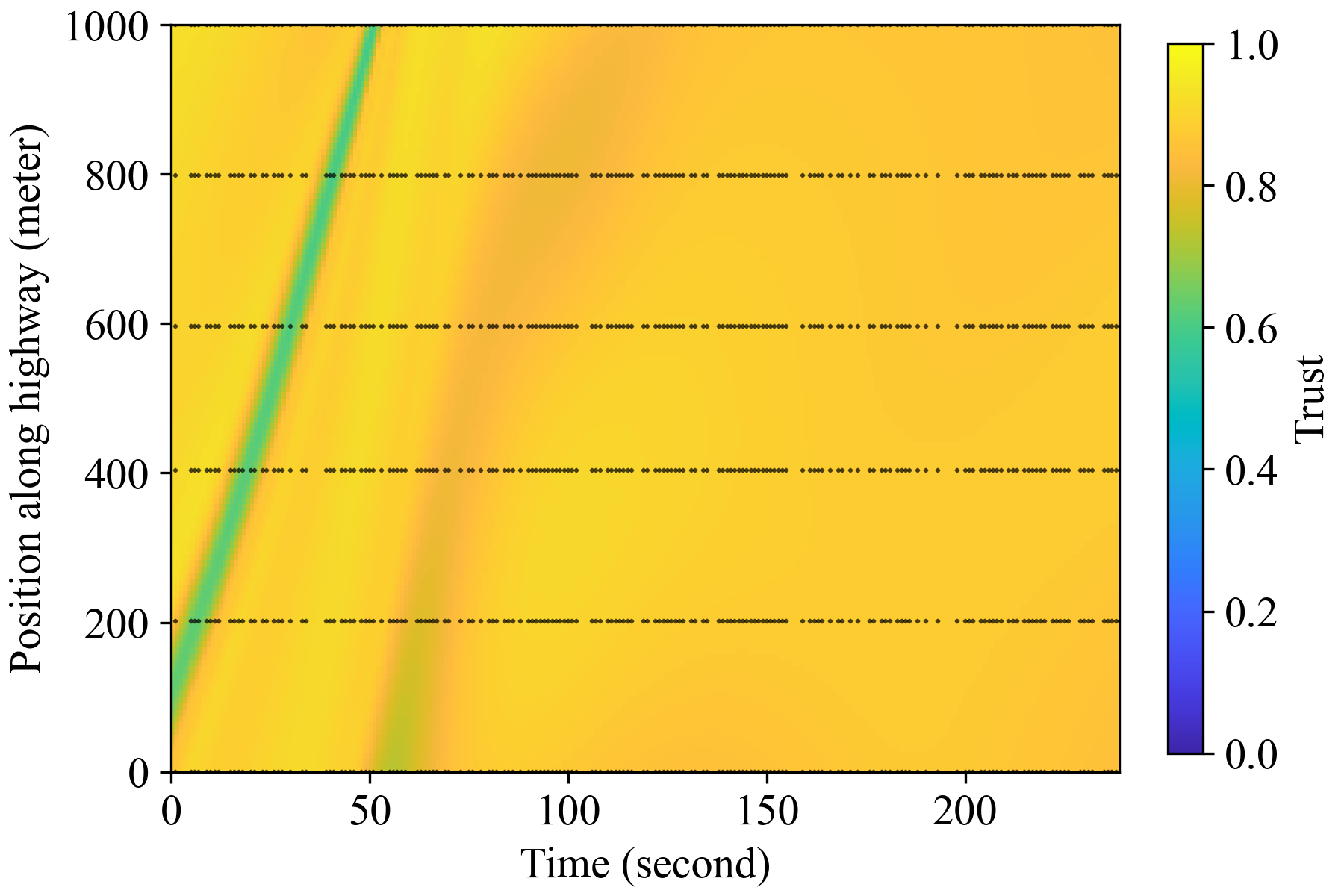}
        \caption{Without field properties}
        \label{fig:impl_dl}
    \end{subfigure}
    \hfill
    \begin{subfigure}[b]{0.32\textwidth}
        \centering
        \includegraphics[width=\textwidth]{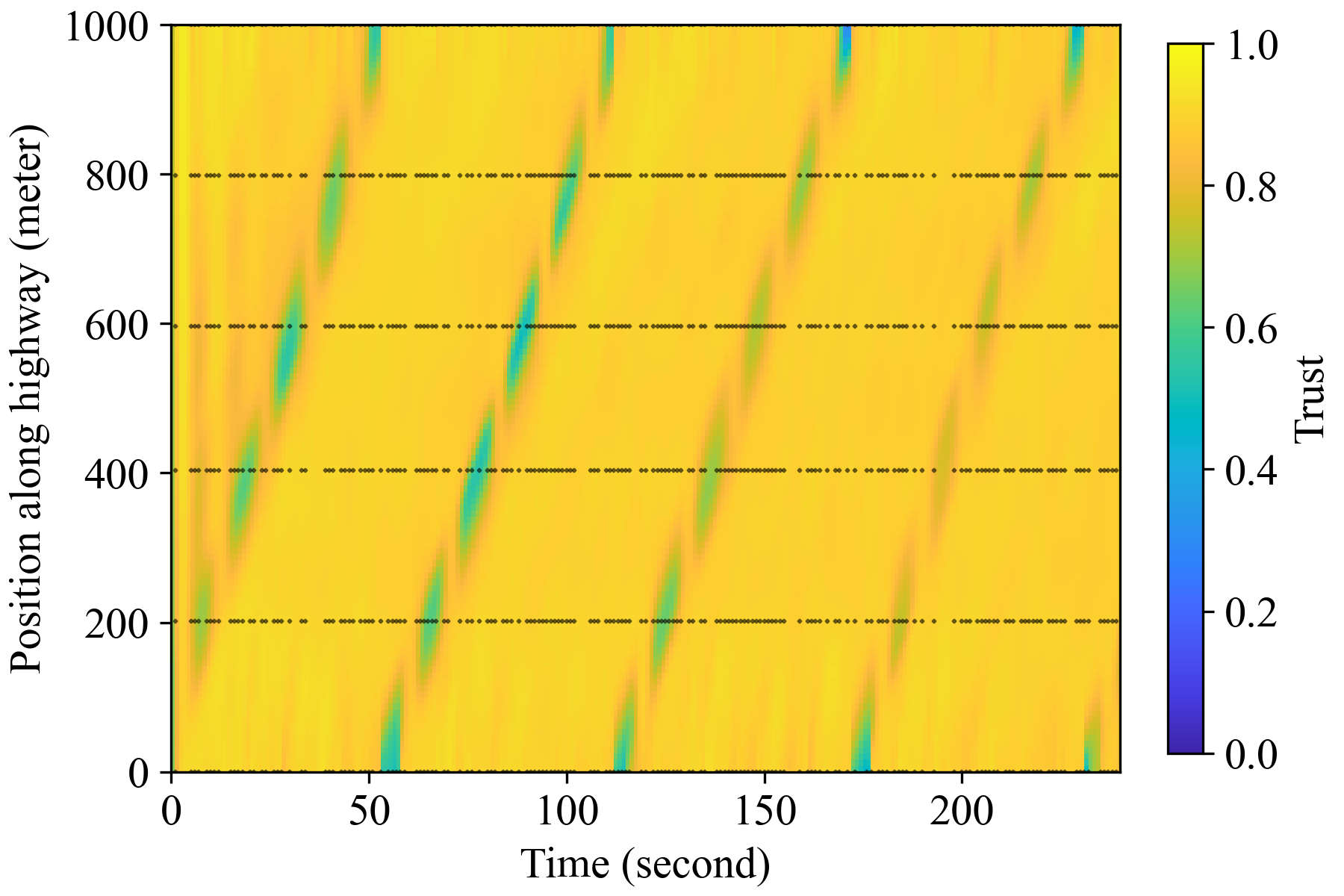}
        \caption{With field properties}
        \label{fig:impl_fidl}
    \end{subfigure}

    \caption{Trust state estimation with and without trust field properties.}
    \label{fig:implication}
\end{figure*}

\subsection{Trust State Estimation Without Trust Field Properties}
\label{subsec:without_tf}

As a reference case, we consider a reconstruction approach that employs deep learning without exploiting field structure or vehicle trajectory information. In this setting, reconstruction relies solely on sparse RSU trust observations $\{Y_{m,k}\}$.

A feedforward neural network, specifically a multilayer perceptron (MLP), is used to approximate the trust field as a function of spatial location and time,
\begin{equation}
\hat T_{\mathrm{DL}}(x,t) = f_\theta(x,t),
\label{eq:dl_field}
\end{equation}
where $f_\theta(\cdot)$ denotes a nonlinear mapping parameterized by $\theta$. The MLP consists of multiple fully connected layers with nonlinear activation functions, enabling it to represent smooth surfaces over the $(x,t)$ domain.

The network is trained by minimizing a data-fitting loss evaluated at the RSU sensing points,
\begin{equation}
\mathcal{L}_{\mathrm{DL}}
=
\sum_{m,k}
\left|
f_\theta(x_m,t_k) - Y_{m,k}
\right|^2.
\label{eq:dl_loss}
\end{equation}

\subsection{Trust State Estimation With Trust Field Properties}
\label{subsec:tf_informed}

In contrast, the field-informed reconstruction explicitly incorporates the generative structure of the trust field and the Lagrangian dynamics of vehicle motion. The objective is to infer latent trust trajectories $\{\tau_i(t)\}$ that are consistent with the observed RSU measurements. Kernel-based weights
\begin{equation}
w_{m,i}(t_k)
=
\frac{K_\sigma\!\left(x_m - x_i(t_k)\right)}
{\sum_j K_\sigma\!\left(x_m - x_j(t_k)\right)}
\label{eq:tidl_weights}
\end{equation}
are used to softly associate RSU trust measurements with individual vehicles, yielding vehicle-aligned trust signals that evolve along each trajectory.

A recurrent neural network (RNN) is then employed to estimate latent trust trajectories over time. In this work, a gated recurrent unit (GRU) architecture is adopted to capture temporal dependencies while maintaining a compact parameterization. The GRU shares parameters across all vehicles and produces trust estimates of the form
\begin{equation}
\hat\tau_i(t) = g_\phi\!\left(\mathbf{z}_i(t)\right),
\end{equation}
where $\mathbf{z}_i(t)$ denotes the input feature vector for vehicle $i$ at time $t$.

Training enforces consistency between observed RSU measurements and those reconstructed via forward aggregation of the estimated vehicle trust,
\begin{equation}
\hat Y_{m,k}
=
\sum_i \hat\tau_i(t_k)\, w_{m,i}(t_k),
\end{equation}
together with regularization terms that encode realistic trust dynamics, including temporal smoothness and a high-trust prior for most vehicles. Once vehicle-level trust trajectories are inferred, the full spatio-temporal trust field is reconstructed using \cref{eq:trust_field}.

\begin{table}[ht!]
\centering
\caption{RMSE and MAE comparison.}
\renewcommand{\arraystretch}{1.3}
\begin{tabular}{|cc|cc|}
\hline
\multicolumn{2}{|c|}{RMSE}                     & \multicolumn{2}{c|}{MAE}                      \\ \hline
\multicolumn{1}{|c|}{Without Info} & With Info & \multicolumn{1}{c|}{Without Info} & With Info \\ \hline
\multicolumn{1}{|c|}{0.055211}     & 0.039285  & \multicolumn{1}{c|}{0.032670}     & 0.021996  \\ \hline
\end{tabular}
\label{tab:rmse_mae}
\end{table}

\subsection{Estimation Results}
\label{subsec:results_implication}

\cref{fig:implication} compares the reconstructed trust fields from both approaches with the ground truth. The true field (\cref{fig:impl_ground}) exhibits clear diagonal low-trust bands propagating in space and time, reflecting the Lagrangian nature of trust carried by malicious vehicles.

The deep learning model without field information (\cref{fig:impl_dl}) fails to recover these coherent structures. Trained purely over Eulerian space–time coordinates, the MLP captures smooth interpolation patterns aligned with sampling geometry rather than vehicle-induced dynamics, resulting in smeared and disconnected low-trust regions.

In contrast, the field-informed approach (\cref{fig:impl_fidl}) explicitly models trust as a latent vehicle-carried quantity and incorporates trajectory information during learning. It successfully reconstructs diagonal low-trust bands with correct orientation, extent, and persistence, closely matching the ground truth.

Quantitative results in \cref{tab:rmse_mae} confirm these findings: without field information, RMSE and MAE are 0.055211 and 0.032670, respectively; incorporating field structure reduces them to 0.039285 and 0.021996, corresponding to relative improvements of about 29\% (RMSE) and 33\% (MAE). These results demonstrate that embedding trust-field structure significantly improves reconstruction accuracy under sparse observations.

\section{Conclusion and Future Work}
\label{sec:conclusion}

This paper introduces a spatio-temporal \emph{trust-field} framework for vehicular ad hoc networks, offering a macroscopic representation of trust analogous to traffic-flow fields. A continuous trust field is defined on road segments using normalized kernel aggregation. Simulation experiments with synthetic trajectories show that microscopic trust perturbations generate structured macroscopic patterns: in free flow, malicious behavior forms a diagonal low-trust band aligned with vehicle trajectories, while in congestion it produces localized, persistent degradations shaped by shockwaves. Trust-field inference from sparse RSU observations further highlights its value: a coordinate-based deep learning baseline fails to recover trajectory-aligned structures, whereas a field-informed model treating trust as a latent vehicle-carried quantity more accurately reconstructs coherent patterns and achieves lower error.

Although this work establishes the theoretical foundations and validation of the trust layer, it does not yet quantify how trust-field dynamics influence the communication layer or the transportation layer. Future work will extend the framework to real-world trajectory datasets under realistic sensing and communication constraints and integrate the trust-field model with communication and traffic simulators to enable closed-loop, cross-layer evaluation of trust-aware control and information-sharing policies.

\bibliographystyle{IEEEtran}
\bibliography{existance_of_trust_field}

\end{document}